\documentclass[10pt,journal,compsoc]{IEEEtran}
\ifCLASSOPTIONcompsoc
  \usepackage[nocompress]{cite}
\else
  \usepackage{cite}
\fi
\ifCLASSINFOpdf
\else
\fi
\usepackage{times}
\usepackage{epsfig}
\usepackage{graphicx}
\usepackage{amsmath}
\usepackage{amssymb}
\usepackage{listings}
\usepackage[english]{babel}
\usepackage{multicol}
\usepackage{xspace}
\usepackage[table,xcdraw]{xcolor}

\makeatletter
\DeclareRobustCommand\onedot{\futurelet\@let@token\@onedot}
\def\@onedot{\ifx\@let@token.\else.\null\fi\xspace}
\def\eg{\emph{e.g}\onedot}

\begin{document}
%
\title{Knowledge-based Embodied Question Answering}
%
%
%
%


\author{Sinan~Tan,
        Mengmeng~Ge,
        Di~Guo,
        Huaping~Liu,
        and~Fuchun~Sun

\IEEEcompsocitemizethanks{
\IEEEcompsocthanksitem Corresponding author: Huaping Liu. hpliu@tsinghua.edu.cn\\
\IEEEcompsocthanksitem Sinan Tan and Mengmeng Ge contribute equally to the paper.\protect\\
\IEEEcompsocthanksitem Sinan Tan, Mengmeng Ge, Di Guo, Huaping Liu and Fuchun Sun are with Tsinghua University.}
\thanks{Manuscript received August 8, 2021.}}

%
%

\markboth{}%
{Shell \MakeLowercase{\textit{et al.}}: Bare Demo of IEEEtran.cls for Computer Society Journals}
%



\IEEEtitleabstractindextext{%
\begin{abstract}
In this paper, we propose a novel \textbf{Knowledge-based Embodied Question Answering (K-EQA)} task, in which the agent intelligently explores the environment to answer various questions with the knowledge. Different from explicitly specifying the target object in the question as existing EQA work, the agent can resort to external knowledge to understand more complicated question such as ``Please tell me what are objects used to cut food in the room?'', in which the agent must know the knowledge such as ``knife is used for cutting food''. 

To address this K-EQA problem, a novel framework based on neural program synthesis reasoning is proposed, where the joint reasoning of the external knowledge and 3D scene graph is performed to realize navigation and question answering. Especially, the 3D scene graph can provide the memory to store the visual information of visited scenes, which significantly improves the efficiency for the multi-turn question answering. Experimental results have demonstrated that the proposed framework is capable of answering more complicated and realistic questions in the embodied environment. The proposed method is also applicable to multi-agent scenarios.
\end{abstract}

}

\maketitle

\IEEEdisplaynontitleabstractindextext

%
\IEEEpeerreviewmaketitle

\IEEEraisesectionheading{\section{Introduction}\label{sec:introduction}}

\begin{table*}
\begin{center}
\begin{tabular}{|l|c|c|c|c|c|c|}
\hline
Method & Navigation + QA & Multi-Target & Logical Reasoning & Interaction & Knowledge & Multi-Turn \\
\hline\hline
EQA\cite{das2018embodied}         & \checkmark & --          & --       &  --          & --           &  --          \\
VideoNavQA\cite{cangea2019videonavqa}  & \checkmark & \checkmark & --           &  --          & --           & --           \\
MT-EQA\cite{yu2019multi}      & \checkmark & \checkmark & \checkmark & --           & --           & --           \\
IQA\cite{gordon2018iqa}      & \checkmark & \checkmark & --           & \checkmark & --           & --           \\
K-EQA(ours)       & \checkmark & \checkmark & \checkmark & --           & \checkmark & \checkmark \\
\hline
\end{tabular}
\end{center}
\caption{A comparison of our proposed K-EQA task with other existing EQA tasks.}
\label{table:Dataset_Comparation}
\end{table*}

\IEEEPARstart{R}{ecently} in AI communities, we have been pursuing the goal of enabling intelligent agents to help humans accomplish practical tasks, such as actively perceiving the environment, searching for objects, and answering questions. As a big step towards useful AI agents, Embodied Question Answering (EQA) \cite{das2018embodied}, a type of task that controls an agent to navigate in a 3D environment and answer a specific problem given by natural language, has been getting more and more attention.

\begin{figure}[t]
\begin{center}
\includegraphics[width=\linewidth]{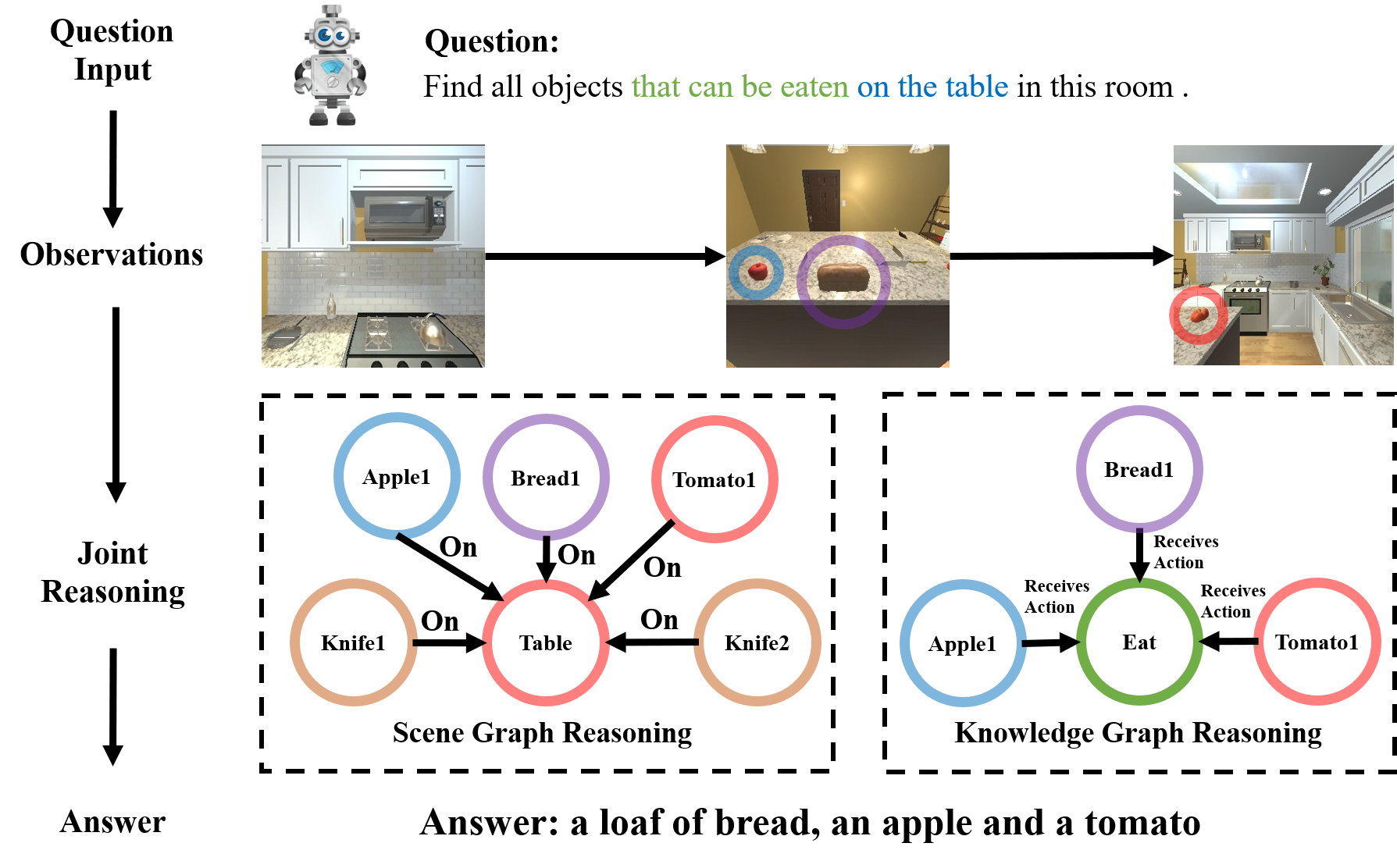}
\end{center}
   \caption{An intuitive demonstration of Knowledge-based EQA. The agent is asked to ``Find all objects that can be eaten on the table in this room.''. To answer this question, the agent needs to both understand the 3D scene, realizing the existence of the apple, the bread and the tomato, and get knowledge from the knowledge graph, learning about the relation triplets (`Apple', `ReceivesAction', `Eat'), (`Bread', `ReceivesAction', `Eat'), (`Tomato', `ReceivesAction', `Eat'). To this end, the exploration and joint reasoning are employed to solve this problem.}
\label{fig:Introduction}
\end{figure}

Currently, most EQA tasks \cite{das2018embodied, gordon2018iqa, yu2019multi} mainly focus on simple questions, in which utilizing information inside the 3D environment is enough to answer the question. However, these EQA tasks and methods have some drawbacks in real-world applications. \textbf{Firstly}, the target object in the question is explicitly specified and the embodied agent can not use external knowledge to answer complicated questions, while such a problem task has been discussed  for vision and language in many recent publications \cite{wang2018fvqa, marino2019ok, wu2017image}. For example, to answer ``Please tell me what are objects used to cut food in the room?'', the agent would have to query the knowledge graph to learn about the relationship (`Knife', `Used To', `Cut food'). Only after learning about this, the agent would know the knife is a target when it comes across one.\textbf{Secondly}, the logical reasoning is not supported, while questions in real-world scenarios could have logical relations. For example, if someone who considers both apple and tomato are acceptable and wants to find something to eat may ask ``Are there any apple \textit{or} tomato in the room?''. This example involves more than one object and the logical relation between objects in the question. Although this scenario is previously discussed in Multi-Target Embodied Question Answering (MT-EQA) \cite{yu2019multi}, it has the limitation that each object exists only once, and it uses a template-based method to break down the questions, which may be powerless to understand sentences with complex structures. \textbf{Thirdly}, the agent is not aware of the already explored parts of a scenario. For example, someone preparing dinner may ask ``Is there any apple on the table?''. And then ask the agent to do something else. When the food is ready, he might then ask ``Are there any objects used for storing food on the table?''. With multiple turns of questions coming to the agent in the same environment, the agent should be able to utilize the previous memories to avoid unnecessary exploration.


To investigate the above problems, we propose a novel Knowledge-based EQA problem and a framework that utilizes the knowledge and 3D scene graph to help the agent navigate in the environment and answer the question (Figure \ref{fig:Introduction}). The proposed method is based on neural-program-synthesis, which translates the question to a Structured Query Language(SQL) statement, enabling the agent to handle questions requiring knowledge and logical reasoning. Additionally, the 3D scene graph can provide the memory to store the visual information of visited scenes,  which significantly improves the efficiency for the multi-turn question answering. Moreover, the proposed framework plans the agent’s motion by selecting relevant sub-regions of the scene and uses a Monte-Carlo tree search (MCTS)-based executor to control its low-level action, making it extendable to multi-agent scenarios easily.

Furthermore, we present the K-EQA dataset --- a new dataset for the K-EQA task with logic clauses and knowledge-related phrases, which will be released soon. We also propose a novel data generation method defined by a formal grammar. This method would generate more complex and various questions in the dataset. Table \ref{table:Dataset_Comparation} compares our proposed dataset with other existing EQA datasets.

\noindent \textbf{Contributions.} Our main contributions include:
\begin{itemize}
\item \textbf{Problem.} We present a generalization of the existing EQA task called K-EQA, which features knowledge and scene graph based reasoning, logical reasoning, holistic scene understanding, and multi-turn \& multi-agent question answering.

\item \textbf{Dataset.} We propose a novel approach to generate the dataset for the K-EQA task, utilizing a formal grammar, scene graph data, and external knowledge base data. The generated questions in the dataset are more enriched and require knowledge and logical reasoning to answer, providing support for more realistic and complicated QA tasks in an Embodied environment.

\item \textbf{Method.} We propose a neural-program-synthesizer-based framework for K-EQA, which generates a SQL program to reason over the 3D scene graph and the knowledge graph. The answering program will be executed on the 3D scene graph and the knowledge graph to answer the question. The planning program plans the agent's motion by selecting relevant sub-regions using the generated SQL program and uses an MCTS-based route planner to visually cover these regions. The framework solves the proposed knowledge-based EQA (K-EQA) problems and works well in multi-turn and multi-agent question answering.

\end{itemize}

\begin{table*}[ht]
\begin{center}
\begin{tabular}{|l|c|c|c|c|c|c|c|c|c|c|c|c|c|c|}
\hline
& \multicolumn{3}{c|}{Existence} & \multicolumn{6}{c|}{Counting} & \multicolumn{3}{c|}{Comparing} & \multicolumn{1}{c|}{Enum.} & \multicolumn{1}{c|}{All} \\ \hline
Split & Yes  & No & Total & 0 & 1 & 2 & 3 & 4 & Total & Yes & No & Total & Total & Total \\
\hline

\hline Train & 6250 & 6250 & \textbf{12500} & 2500 & 2500 & 2500 & 2500 & 2500 & \textbf{12500} & 6250 & 6250 & \textbf{12500} & \textbf{12500} & \textbf{50k} \\
\hline Test & 1250 & 1250 & \textbf{2500} & 500 & 500 & 500 & 500 & 500 & \textbf{2500} & 1250 & 1250 & \textbf{2500} & \textbf{2500} & \textbf{10k} \\
\hline

\end{tabular}
\end{center}
\caption{Question types and answers distribution of the proposed K-EQA dataset, applied to both K-EQA and K-EQA Extension. ``Enum.'' stands for ``Enumerating''.}
\label{table:Question_Distribution}
\end{table*}

\section{Related Work}

\subsection{Embodied Question Answering}

Our work is a generalization of the recently proposed Embodied Question Answering (EQA) \cite{das2018embodied, wijmans2019embodied, chen2019audio, luo2019segeqa} task, where the agent needs to explore the scene and answer a given question. After EQA is proposed, many variants are developed to improve the logical reasoning of the given questions. Multi-Target Embodied Question Answering (MT-EQA) \cite{yu2019multi} uses a template-based method to expand the original EQA from the single-target setting to a challenging multi-target setting, which requires the agent to perform comparative reasoning before executing answering. Neural modular control \cite{das2018neural} applies neural module networks \cite{andreas2016neural} to EQA tasks. Interactive Question Answering (IQA) \cite{gordon2018iqa, gordon2019should} considers cases where objects of interest are distributed in many different locations. Recent works \cite{jain2019two, das2019tarmac, tanmulti} focus on multi-agent cooperation in embodied environments.


\subsection{Question Answering with Knowledge and Logic}

Traditional Visual Question Answering (VQA) \cite{antol2015vqa} datasets focus on questions answered by only direct analysis of image and question. These questions require limited common sense and factual knowledge to answer. Recently, several works \cite{marino2019ok, wang2018fvqa} start to include external knowledge and facts to the VQA datasets. \cite{liang2019focal} proposes MemexQA task where the answer contains not only text but also photos of history events. \cite{liu2018inverse} proposes inverse VQA tasks where question needs to be generated according to the image and the answer. These works use scene graph representation and grammars to generate complex questions. \cite{das2017visual} proposes visual dialog dataset requiring dialog context reasoning. \cite{wang2018fvqa} composes questions requiring external knowledge by introducing relations triplets in knowledge bases like ConceptNet to the questions. We following a similar approach to generate knowledge-based reasoning clauses in our dataset generation process.

To investigate and tackle the limit of existing VQA models in complex scenarios with questions requiring logical reasoning. CLEVR \cite{johnson2017clevr, kottur2019clevr} and GQA \cite{hudson2019gqa} are proposed to diagnose visual reasoning abilities on VQA tasks. These datasets have inspired several methods \cite{andreas2016neural, xiong2016dynamic, hu2018explainable, vedantam2019probabilistic, cao2019interpretable} emphasizing visual reasoning ability.



\section{Problem Formulation}
The Knowledge-based EQA is formulated as a question-answering task in an embodied environment $\mathcal{E}$ with knowledge graph $\mathcal{G}_K$. At each time step $t$, the agent utilizes the knowledge graph $\mathcal{G}_K$, the question $Q$, and the previous observations $o_{1:t-1} = (o_1, o_2, ..., o_{t-1})$ to generate the action as:

\begin{equation*}
    {a_t} = \pi_{nav}(\mathcal{G}_K, Q, o_{1:t-1})
\end{equation*}

After performing the action $a_t$, the agent receives a new observation $o_t$ from $\mathcal{E}$. Such a procedure iterates until the agent triggers a special stop action at time instant $T$. Then the agent could get the answer from

\begin{equation*}
    {ans} = \pi_{ans}(\mathcal{G}_K, Q, o_{1:T-1})
\end{equation*}

The goal is to design appropriate policies $\pi_{nav}$ and $\pi_{ans}$ which leverages the advantages of the training data and the external knowledge. This problem significantly differs from existing work such as \cite{gordon2018iqa, das2018embodied, yu2019multi} because of the introduction of the knowledge graph, which provides strong support for navigation in complex environments and answering complicated problems.

\section{K-EQA Dataset}
\definecolor{morange}{RGB}{224,130,68}
\definecolor{mred}{RGB}{235,50,35}
\definecolor{mblue}{RGB}{76,116,190}
\definecolor{mgreen}{RGB}{127,171,85}
\begin{table*}[t]
\centering
\begin{tabular}{|l|l|l|}
\hline 
	& \multicolumn{1}{c|}{K-EQA} & \multicolumn{1}{c|}{K-EQA Extension} \\ 
\hline
Existence & \begin{tabular}[c]{@{}l@{}}Q: Is there \textcolor{mred}{an object} \textcolor{mgreen}{used to cut food} \textcolor{mblue}{near a} \\ \textcolor{mblue}{salt shaker} somewhere nearby?\\ A: Yes \end{tabular}
& \begin{tabular}[c]{@{}l@{}}Q: Is there \textcolor{mred}{an object} \textcolor{mgreen}{that is a type of herb} \textcolor{mblue}{near an egg} \textcolor{morange}{or} \textcolor{mred}{a} \\\textcolor{mred}{roll} \textcolor{mred}{of paper towel} \textcolor{mblue}{near an egg} in the room?\\ A: Yes\end{tabular}\\ 
\hline
Counting  & \begin{tabular}[c]{@{}l@{}}Q: Please tell me how many \textcolor{mred}{objects} \textcolor{mgreen}{which} \\ \textcolor{mgreen}{you can use to hold rice} are around here?\\ A: 2\end{tabular}   
& \begin{tabular}[c]{@{}l@{}}Q: How many \textcolor{mred}{pens} \textcolor{mblue}{near a desk} \textcolor{morange}{or} \textcolor{mred}{objects} \textcolor{mgreen}{used to listen to} \\ \textcolor{mgreen}{music} \textcolor{mblue}{above an alarm clock} \textcolor{morange}{or} \textcolor{mred}{books} \textcolor{mblue}{above an alarm clock}\\are there in the room?\\ A: 3\end{tabular} \\ 
\hline
Comparing & \begin{tabular}[c]{@{}l@{}}Q: Are there more \textcolor{mred}{spray bottles} than \textcolor{mred}{objects}\\ \textcolor{mgreen}{for drying the body after a bath} \textcolor{mblue}{near a toilet} \\in the room?\\ A: No\end{tabular} 
& \begin{tabular}[c]{@{}l@{}}Q: Are there less \textcolor{mred}{objects} \textcolor{mgreen}{used to store binary information} \textcolor{morange}{or}\\ \textcolor{mred}{credit cards} \textcolor{mblue}{above a keychain} than \textcolor{mred}{objects} \textcolor{mgreen}{that are a kind of}\\ \textcolor{mgreen}{mechanical device} \textcolor{morange}{or} \textcolor{mred}{cell phones} in the room ?\\ A: Yes\end{tabular} \\ \hline
Enumerating & \begin{tabular}[c]{@{}l@{}}Q: there are some \textcolor{mred}{objects} \textcolor{mgreen}{for eating} \textcolor{mblue}{in the} \\ \textcolor{mblue}{room}. What are they?\\ A: A loaf of bread, a tomato and 2 apples\end{tabular}             
& \begin{tabular}[c]{@{}l@{}}Q: There are some \textcolor{mred}{objects} \textcolor{mgreen}{which are used for sleeping} \textcolor{morange}{or}\\ \textcolor{mred}{objects} \textcolor{mgreen}{for cuddling} in the room. What are they?\\ A: A pillow and a teddy bear\end{tabular}\\ 
\hline
\end{tabular}
\caption{Sample questions and answers of each type in K-EQA and K-EQA Extension. Object category specifiers are colored in red; knowledge graph clauses are colored in green; spatial relations are colored in blue; logical connectors are colored in orange.}
\label{table:Question_Samples}
\end{table*}


Our dataset is built upon AI2Thor \cite{ai2thor}, a photorealistic 3D environment for embodied vision research. The AI2Thor simulator contains 120 different room layouts of 4 categories (Bedroom, Living room, Kitchen, and Bathroom), with 30 room layouts for each category. Each layout allows randomizing the quantities and locations of all ``pickupable'' objects (\eg Apple, Knife, and Book). The proposed dataset is split into 2 subsets --- \textbf{K-EQA} and \textbf{K-EQA Extension}, where the \textbf{K-EQA} split contains questions more likely to appear in daily life, while the \textbf{K-EQA Extension} split contains more complex questions challenging the reasoning ability of the model, which could be useful for challenging the limit of reasoning ability for the methods. For both subsets, the dataset organization is similar --- We create 50 scenes for each layout in AI2Thor, yielding 6,000 different rooms in total. With 10 questions are asked to the agent for each room, the generated dataset has 60,000 questions in total. For each room category, 25 layouts will be used for training and the other 5 will be used for testing. Therefore the training split will have 50,000 questions, and the test split will have 10,000 questions. 

\subsection{Choosing Entities and Relations for Generating Questions}

In the first step of building the dataset, we will map each entity that appeared in the AI2Thor environment \cite{ai2thor} to a set of knowledge base relations to choose from. We select ConceptNet \cite{speer2016conceptnet} as the knowledge base used in our work, which contains more than 1.5M entities for English Language. To choose possible relation triplets for our dataset, we first restrict all the relations in the knowledge base to the ones including entities directly or indirectly related to objects appearing in the AI2Thor dataset. (Here ``indirectly'' means that some knowledge would require multi-step reasoning, for example, the following relation triplet  (`Basketball', `ReceivesAction', `purchased at a sporting goods store') could be inferred from the following relations in ConceptNet: (`Basketball', `IsA', `Basketball equipment'),  (`Basketball equipment', `IsA', `Sports equipment') and (`Sports equipment', `ReceivesAction', `purchased at a sporting goods store')).

We then manually annotate some entities to further filter some misformed relations from the knowledge base to improve the quality of our dataset. The final set of relations used for question-answering in our work contains 7,451 relation triplets.

\subsection{Question Types and the Dataset}


There are 4 different types of questions in our dataset: Existence, Counting, Enumerating and Comparing. The ``Existence'' problem and ``Comparing'' problems ask whether objects meet certain condition exists, or are more/less than objects of another category. The ``Counting'' problem asks the number of objects meeting certain conditions. The most difficult type of problem is the ``Enumerating'' problem, which asks about the exact number of objects by category. Table \ref{table:Question_Samples} shows examples of questions and their answers.

To generate questions using the relations we have chosen from the knowledge base, we extend the templates provided in IQA and MT-EQA to a set of grammar. For each type, objects can be specified directly by their category, or by relations with other objects in the scene, or by relations with other entities in the knowledge graph. These object type specifiers could be connected with logic connector ``\textit{and}'' or ``\textit{or}'' in K-EQA Extension, challenging the reasoning ability of the question answering model. The proposed method for generating questions is similar to \cite{johnson2017clevr, hudson2019gqa}.

\subsection{Question generation}
Now we are looking into the details of how exactly a question is generated. A question in the proposed dataset is generated using the following steps. Note that the following steps apply to \textbf{KEQA Extension}. For the basic \textbf{K-EQA} dataset, some steps will be skipped:
\begin{itemize}
    \item \textbf{Question type sampling}: A question type among the 4 types of questions (Counting, Existence, Comparing, Enumerating) is chosen. We generate 300 questions of each type for each scene so that there will be enough questions for further steps regarding dataset balancing.
    \item \textbf{Object count sampling}: The number of object filters (i.e. the element in the grammar that would choose an object category, and knowledge graph restriction or a scene graph restriction) that will appear in the question will be randomly chosen. For Comparing problems, there are 2 groups of object filters, and the number of object filters for each group will be randomly chosen respectively.
    \item \textbf{Logic connector sampling}. For questions with the type "Existence", the logic connector could be ``and'' or ``or''. And for Comparing Problem, the compare word (could be ``more'' or `` less'') will also be randomly chosen. This step only applies to \textbf{KEQA Extension}.
    \item \textbf{Object Sampling}: Objects are chosen randomly from the list of all possible appearing objects for a given room type. Note that for a specific room, the chosen object may exist or not exist.
    \item \textbf{Introducing knowledge graph}: A precomputed object-to-entity and entity-to-relation mapping are used. The question generator will randomly decide whether or not to use knowledge to refer to an object category. If it decides to use knowledge to refer to that object, the category will be replaced by entity and relations in the knowledge graph.
    \item \textbf{Adding scene graph relations}: For each possible object, a scene relation could be randomly added to the object. Similar to the knowledge graph relations, all possible scene graph relations are also precomputed using the ground truth scene graph.
    \item \textbf{Computing the ground truth answer}: The answer is computed using the ground truth scene graph, the syntax tree, and the knowledge graph.
\end{itemize}

\subsection{Balancing and Generating Datasets}
As mentioned above, the objects in our dataset may be specified by their categories, relations with other objects, or relations with other entities in the knowledge graph. With these low-level language features, it would be important to eliminate potential biases in the answer distribution. To make the proposed dataset balanced, we will first create a ``tag'' for each question to represent these low-level language features. For example, a question, with the type ``Comparing'', comparing word ``less'' and spatial relations ``near'' and ``above'', will be tagged as ``COMPARE\_less\_1\_1\_No\_near\_above''. We will firstly generate a huge number of questions, resulting in a question sampling pool. Then we would remove some questions from the sampling pool so that each kind of tag brings no biased distribution of answers in the dataset. To be specific, we use a 2-stage algorithm to balance the question pool for our proposed dataset: 

\subsubsection{Balancing the proposed dataset}
To make a balanced dataset, we first generate more questions than we need to build the dataset, and then we would assign a tag for each question. The question format for each kind of question would be:
\begin{itemize}
    \item \textbf{Counting}: Tag format consists of ``COUNTING'', number of object filters, and the answer.
    \item \textbf{Existence}: Tag format consists of ``EXISTENCE'', logic Connector, number of object filters, and the answer.
    \item \textbf{Comparing}: Tag format consists of ``COMPARE'', compare Word, number of object filters in the first group, number of object filters in the second group, and the answer.
    \item \textbf{Enumerating}: Tag format consists of ``ENUMERATING'', number of object filters, and number of objects in the answer.
\end{itemize}

When sampling from the generated questions to form the proposed dataset, we will control the exact number of questions belonging to each tag of each type. This makes our proposed dataset precisely balanced in terms of answer and object distribution, as is shown in Table \ref{table:Question_Distribution}.

\subsubsection{Sub-tag balancing}
However, the objects in our dataset may be specified by their categories, relations with other objects, or relations with other entities in the knowledge graph. Making the tag-level balancing inadequate to provide a balanced dataset. To eliminate potential biases in the answer distribution, we will first create a ``sub-tag'' for each question to represent these low-level language features. For example, a question, with the type ``Comparing'', comparing word ``less'', and each part to compare containing only 1 object specifier, and spatial relations ``near'' and ``above'', and the answer ``No'', will be tagged as ``COMPARE\_less\_1\_1\_No\_near\_above''. 
Despite we have already generated a large number of questions for a scene, making the dataset precisely balanced with regards to sub-tag is impossible. Therefore we would remove some questions from the sampling pool so that each kind of tag brings no biased distribution of answers in the dataset. (e.g. the number of questions with tag ``COMPARE\_less\_1\_1\_No\_near\_above'' and the number of questions with tag ``COMPARE\_less\_1\_1\_Yes\_near\_above'' should be the same.). Despite that the number of questions will not be balanced for different types of sub-tags, the answer distribution for each sub-tag is a uniform one.

Because of the sub-tag balancing technique, when sampling from the generated questions, the answer distribution for every kind of sub-tag is balanced. Despite the number of questions of each sub-tag is not the same, it would be not possible to infer the answer from those low-level language features.

All possible tags of the proposed dataset are provided in Table \ref{table:Question_Tags}.

\begin{table*}[h]
\centering
\begin{tabular}{|l|c|c|c|c|}
\hline 
	& \multicolumn{1}{c|}{Existence} & \multicolumn{1}{c|}{Counting} & \multicolumn{1}{c|}{Comparing} & \multicolumn{1}{c|}{Enumerating} \\ 
\hline 
 K-EQA  & \begin{tabular}[c]{@{}l@{}} EXISTENCE\_1\_No \\ EXISTENCE\_1\_Yes \end{tabular} & \begin{tabular}[c]{@{}l@{}}COUNTING\_1\_0 \\ COUNTING\_1\_1 \\ COUNTING\_1\_2 \\ COUNTING\_1\_3 \\ COUNTING\_1\_4\end{tabular} & \begin{tabular}[c]{@{}l@{}}COMPARE\_less\_1\_1\_No \\ COMPARE\_less\_1\_1\_Yes\\ COMPARE\_more\_1\_1\_No \\ COMPARE\_more\_1\_1\_Yes \end{tabular} & \begin{tabular}[c]{@{}l@{}}ENUMERATING\_1\_1 \\ ENUMERATING\_1\_2 \\ ENUMERATING\_1\_3 \\ ENUMERATING\_1\_4\end{tabular}\\
\hline
\hline
 K-EQA Extension	& \begin{tabular}[c]{@{}l@{}}COUNTING\_2\_0 \\ COUNTING\_2\_1 \\ COUNTING\_2\_2 \\ COUNTING\_2\_3 \\ COUNTING\_2\_4 \\ COUNTING\_3\_0 \\ COUNTING\_3\_1 \\ COUNTING\_3\_2 \\ COUNTING\_3\_3 \\ COUNTING\_3\_4\end{tabular} &  \begin{tabular}[c]{@{}l@{}} COMPARE\_less\_1\_2\_No \\ COMPARE\_less\_1\_2\_Yes \\ COMPARE\_less\_2\_1\_No \\ COMPARE\_less\_2\_1\_Yes \\ COMPARE\_less\_2\_2\_No \\ COMPARE\_less\_2\_2\_Yes  \\ COMPARE\_more\_1\_2\_No \\ COMPARE\_more\_1\_2\_Yes \\ COMPARE\_more\_2\_1\_No \\COMPARE\_more\_2\_1\_Yes \\ COMPARE\_more\_2\_2\_No \\ COMPARE\_more\_2\_2\_Yes \end{tabular} & \begin{tabular}[c]{@{}l@{}} COMPARE\_less\_1\_2\_No \\ COMPARE\_less\_1\_2\_Yes \\ COMPARE\_less\_2\_1\_No \\ COMPARE\_less\_2\_1\_Yes \\ COMPARE\_less\_2\_2\_No \\ COMPARE\_less\_2\_2\_Yes  \\ COMPARE\_more\_1\_2\_No \\ COMPARE\_more\_1\_2\_Yes \\ COMPARE\_more\_2\_1\_No \\COMPARE\_more\_2\_1\_Yes \\ COMPARE\_more\_2\_2\_No \\ COMPARE\_more\_2\_2\_Yes \end{tabular} & \begin{tabular}[c]{@{}l@{}}ENUMERATING\_2\_1 \\ ENUMERATING\_2\_2 \\ ENUMERATING\_2\_3 \\ ENUMERATING\_2\_4\\ENUMERATING\_3\_1 \\ ENUMERATING\_3\_2 \\ ENUMERATING\_3\_3 \\ ENUMERATING\_3\_4\end{tabular}\\
\hline
\end{tabular}
\caption{Question tags of each type in K-EQA and K-EQA Extension.}
\label{table:Question_Tags}
\end{table*}

We then sample questions from the tag-balanced sampling pool, with constraints controlling exact answer distribution for each question type. As a result of this process, both splits of the proposed K-EQA dataset consist of 60000 questions across 6000 different environment setups, with balanced question types and answer distributions. Table \ref{table:Question_Distribution} shows the dataset splits and the distribution of question types and answers for each split.


\section{Overview of the Proposed Framework}

\begin{figure*}[t]
\begin{center}
\includegraphics[width=\linewidth]{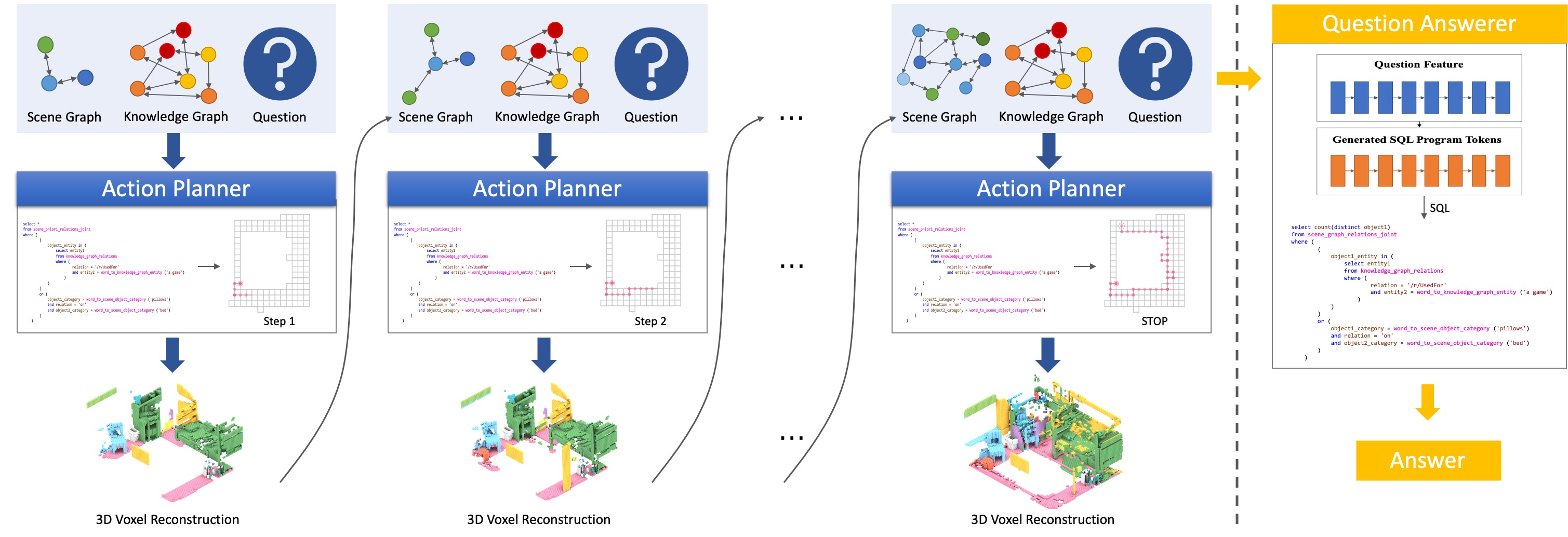}
\end{center}
   \caption{Overview of the proposed framework. The \textbf{Action Planner} uses text-to-SQL translation to select a set of relevant regions and a scan executor to plan the agent's action sequences. 3D scene graph will be updated based on the observations and the 3D voxel reconstruction. When the \textbf{Action Planner} decides there are no more regions to explore, the agent will stop and use the \textbf{Question Answerer} to answer the question.}
\label{fig:framework}
\end{figure*}

As is demonstrated in Figure \ref{fig:framework}, our proposed framework is based on the 3D reconstruction and the 3D scene graph of the environment. As the agent explores, it will maintain a state memory $s_t$ for each step $t$, which is a 3D scene graph built gradually based on the voxel-level reconstruction.

The state memory is used for two purposes: (1) Guide the agent. This is done with the help of the \textbf{Action Planner}, which will first select a set of relevant regions in the scene graph according to the question. A scan executor will plan the agent's low-level actions (e.g. Move/Rotate), according to a map built on the 3D reconstruction and the relevant regions. The agent will then update the state memory according to its observations. (2) Answer the question. When the \textbf{Action Planner} decides there are no longer relevant regions in the 3D environment, the agent will stop and use the \textbf{Question Answerer} to answer the question.

Here are the components of our proposed framework:
\begin{itemize}
    \item \textbf{Scene Graph-based State Memory}: a memory storing visual information perceived in the scene. In our work, the state memory is a 3D scene graph, built upon the environment's voxel-level reconstruction.
    \item \textbf{Action Planner}: The planner first determines whether or not the agent needs to continue the exploration process and which region will be searched when performing exploration. In practice, the Action Planner uses an attentional LSTM-based program synthesizer, taking the question as its input, and generates a SQL program to choose the relevant regions in the current scene graph. After deciding the voxel sets to explore, an MCTS-based scan executor capable of multi-agent controlling will plan the agent's navigational path.
    \item \textbf{Question Answerer}: If the planner decides to stop exploration, the answerer will give the answer based on the scene graph and the knowledge graph. In our implementation, the Question Answerer also generates a SQL program from the question. The program will be able to run on the 3D Scene Graph and the Knowledge Graph, extracting the answer to the given question.
\end{itemize}

\section{Methods and Models}

\subsection{Scene Data Structure}
An important sub-task in our framework is to encode the visual observations to the internal state $s_t$. Here we use a mixture of the voxel-based 3D reconstruction and the 3D scene graph.

\textbf{Voxels.} For each observed image frame, we first run an instance segmentation model (Mask R-CNN \cite{he2017mask}), which will assign a semantic label for each pixel in the image. Then pixels are back-projected into the 3D space with the depth data, forming the voxel-level reconstruction. For each voxel, its 3D coordinate, semantic category, and a confidence score for each category are recorded.

\textbf{Connecting voxels to generate 3D object detections.} Voxels of the same category with a threshold-bounded distance are considered connected. After the connectivity of the voxels is generated, we will be able to get a reconstructed voxel volume, with voxels also form an undirected graph, whose edge is defined by the connectivity of these voxels. The 3D object regions of the scene graph are generated by merging all connected components of the voxel graphs. The maximum coordinate and minimum coordinate of each dimension of the bounding box of these 3D objects will also be determined by the voxel with maximum and minimum coordinate at that dimension.

\textbf{Object-object relation and 3D scene graph generation.} To complete the scene graph, spatial relations need to be generated for all objects. Similar to \cite{armeni20193d}, Here we consider 3 categories of object-object relations in the scene graph:\textbf{(1) Contain, In, Holding, and On} If object A's bounding box overlaps with object B's bounding box, and object B is of a container category, then object B is considered to ``Contain'' object A, and object A is ``Inside'' object B. The ``Holding'' and ``On'' relations are defined similarly, but for objects with plains that other objects could be on (e.g table).\textbf{(2) Above and Below} If the rectangle of the top-bottom view of object A and object B overlaps or their distance is within 0.25m, and the maximum z-coordinate of object B is lower than the minimum coordinate of Object A, then we consider object A is ``Above'' object B, and object B is ``Below'' object A. \textbf{(3) Near} If the distance between the 3D box of Object A and Object B is less than 0.25m. This relation is symmetry.

\subsection{Question Answering}

As was previously specified, our framework requires a question answerer $\pi_{ans}(s_t, Q,\mathcal{G}_K)$. Since now we're using the scene graph $\mathcal{G}_S$ as the state memory, the QA model will need to be graph-based. Here we consider a neural-program-synthesize-based approach: The question will be first converted to a query language statement $\mathcal{P}_{ans}$, then the database software will execute the query language statement as a program on both the scene graph and the knowledge graph. Here we are using SQL as the query language, the Text-to-SQL conversion is a well-discussed topic in the field of natural language processing \cite{yu2018typesql, zhong2017seq2sql, dong2016language}. In our work, we use a synthetic dataset generated from the question grammar introduced in section 4.


We need to generate a program $\mathcal{P}_{ans}$ used to perform the query on both graphs. We formulate the problem as 

\begin{equation}
    \mathcal{P}_{ans} = m_a(Q_1, Q_2, ..., Q_n) 
\end{equation}
where $\mathcal{P}_{ans} = ({P}_{{ans},1}, {P}_{{ans},2}, ..., {P}_{{ans},m})$ is the program, ${P}_{{ans},i}$ is a token in the program. $m_a$ is the translation model, and $Q_i$ is the token of the question. This sub-problem is a sequence-to-sequence translation problem and we use the attention model in \cite{bahdanau2014neural} to solve it.

\subsubsection{Executing Generated Programs}

To make use of the SQL to perform queries on the knowledge graph and the scene graph, we first convert all the scene graph and knowledge graph data to tables, so that the SQL server would be able to execute queries on them. Each entity in the knowledge graph will be added to an entity table, and each object in the scene graph will be added to an object table. These Object-Object relations and Entity-Entity relations are written into other tables, connected to these two tables by foreign keys. The SQL server will then execute the program on the graphs to answer the question. We create custom SQL functions to avoid the program being too long. We use PostgreSQL \cite{postgresql1996postgresql} as the SQL server.

\subsection{Action Planner}
Unlike traditional planners that would control the agent directly, our Action Planner will first specify a range of relevant regions (voxel sets) instead. Similar to the \textbf{Question Answerer}, we need to generate a program $\mathcal{P}_{plan}$. Unlike ${P}_{ans}$, the program ${P}_{plan}$ is not only executed on scene graph and external knowledge graph but also the scene priors extracted from the train set \cite{yang2018visual}. To be specific, the output SQL of the Action Planner generates possible relation triplets that the object relevant to the current question may appear from the scene priors and the external knowledge graph. We then remove all voxels in these irrelevant regions from the regions passed to the scan executor. (For example, for a question searching for basketball, if the relation triplet (``Basketball'', ``above'', ``Bed'') does not appear in the returned query result of the SQL statement, then all voxels above a bed will be excluded from the relevant regions.)

The MCTS-based scan executor use the voxel reconstruction data to determine the visible voxels for each viewport $v = (x, y, \theta, \phi)$. (where $(x, y)$ is the location, and $\theta, \phi$ are the azimuth and inclination angle respectively) Then, the path planning problem becomes finding an optimal viewport sequence $ v^* = (v_1^*, v_2^*, ..., v_n^*)$ so that all voxels in relevant regions can be observed, and the distance of the sequence could be as short as possible.

Similar to previous works \cite{coulom2006efficient, sukkar2019multi}, we use an MCTS-based approach to find the approximate optimal solution to the problem of planning the viewport sequence. The child of each node in the search tree stands for selecting the next view the agent would approach. The value function of each node is set as $1 - \frac {l - l_{min}}{l_{max} - l_{min}}$, where $l$ is the average simulation result (track length) , and $l_{min}, l_{max}$ are global best and worst simulation result, respectively. For multi-agent cases, the agent would need to select the next viewport in turn.

\begin{table*}[h]
\begin{center}
\begin{tabular}{|l|c|c|c|c|c|c|c|c|c|c|}
\hline
& \multicolumn{2}{c|}{Existence} & \multicolumn{2}{c|}{Counting} & \multicolumn{2}{c|}{Comparing} & \multicolumn{2}{c|}{Enum.} & \multicolumn{2}{c|}{Overall} \\ \hline
Dataset & Acc.  & Length & Acc. & Length & Acc. & Length & Acc. & Length & Acc. & Length \\
\hline

\hline K-EQA & $58.8 \%$ & 58.0 & $31.6 \%$ & 87.5 & $62.0 \%$ & 97.7 & $12.8 \%$ & 87.8 & $41.3 \%$ & 82.7 \\
\hline K-EQA Extension & $58.0 \%$ & 81.5 & $29.2 \%$ & 88.1 & $54.4 \%$ & 101.9 & $6.8 \%$ & 95.6 & $37.1 \%$ & 91.8 \\
\hline

\end{tabular}
\end{center}
\caption{Comparing \textbf{K-EQA Extension} experiment results with \textbf{K-EQA} experiment results.  ``Enum.'' stands for ``Enumerating''. ``Acc.'' stands for ``Accuracy''.}
\label{table:Logic_vs_Basic}
\end{table*}

\subsubsection{Details of the action planner and the Executor.}
Similar to the answering SQL, the program generated by the action planner $\mathcal{P}_{plan}$ will look up the SQL tables and return relations triplets. Here we will further explain what the returned relation triplets will be like, using the example ``How many objects used for a game or pillows on a bed are there in the room?''. The planning SQL is:

\newcommand{\verbatimfont}[1]{\renewcommand{\verbatim@font}{\ttfamily#1}}
\verbatimfont{\tiny}
\begin{verbatim}
select *
from scene_priori_relations_joint
where (
        (
            object1_entity in (
                select entity1
                from knowledge_graph_relations
                where (
                        relation = '/r/UsedFor'
                        and entity2 = word_to_knowledge_graph_entity ('a game')
                    )
            )
        )
        or (
            object1_category = word_to_scene_object_category ('pillows')
            and relation = 'on'
            and object2_category = word_to_scene_object_category ('bed')
        )
    )
\end{verbatim}

A part of the typical result returned from such a query of the planner statement will be like the following table:

\begin{table}[h]
\begin{tabular}{|l|l|l|}
\hline
Object1\_Category & Relation & Object2\_Category \\ \hline
Basketball        & Near     & Bed               \\ \hline
Pillow            & On       & Bed               \\ \hline
...               & ...      & Other Objects     \\ \hline
\end{tabular}
\caption{An example of possible partial output of the question ``How many objects used for a game or pillows on a bed are there in the room?''}
\label{table:Planner_Samples}
\end{table}
The explanation for such an output is that basketball will not appear on/above bed (in terms of both common sense and the proposed dataset). ("BasketBall", "Bed", "near") is the only relation that Basketball could have with  (There is also no ("BasketBall", "Bed", "below")) is because Bed often stands the lowest part of the room). If we can already locate regions representing a bed, we will only need to explore regions ``On'' and ``Near'' that bed.

Similar to object relations are defined in Section 6.1, each voxel can be considered to have similar relations to an object. (e.g. ``above'' means that the z-coordinate of that voxel is more than 0.25m higher than the object and the distance of the voxel to the object in the XY-plane is within 0.25m. ). Since now we know that all relevant objects cannot appear on the region ``above'' a bed, these regions will be opted out for the scan executor.

\begin{table}[h]
    \begin{center}
    \begin{tabular}{|l|c|c|c|c|c|}
    \hline
    Method & Exist. & Count. & Cmp. & Enum. & All\\
    \hline
    \hline
    Priori & $50.0\%$ & $20.0\%$ & $50.0\%$ & $1.72\%$ & $30.4\%$ \\
    \hline
    Blind. & $53.0 \%$ &  $19.7 \%$ & $51.3\%$ & $3.5\%$ & $31.9\%$\\
    \hline
    IQA* &49.8\%& 21.4\%& N/A & N/A & N/A\\
    \hline
    w/o KG & $52.4 \%$ & $22.0 \%$ & $49.2 \%$ & $3.6 \%$  & $31.8\%$  \\
    \hline
    Ours & $\textbf{58.8\%}$ & $\textbf{31.6\%}$ & $\textbf{62.0\%}$ & $\textbf{12.8\%}$ &$\textbf{41.3\%}$ \\
    \hline 
    
    \end{tabular}
    \end{center}
    \caption{\textbf{K-EQA} experiment results comparing with \textbf{baseline methods}. ``Cmp.'' stands for ``Comparing'', ``Enum.'' stands for ``Enumerating''. ``Blind.'' stands for ``Blindfold'' \cite{anand2018blindfold}. * For the IQA baseline, we only choose questions considered in both ours and IQA's method \cite{gordon2018iqa}.}
    \label{table:Baseline}
\end{table}

\begin{table*}[h]
\begin{center}
\begin{tabular}{|l|c|c|c|c|c|c|c|c|c|c|}
\hline
& \multicolumn{2}{c|}{Existence} & \multicolumn{2}{c|}{Counting} & \multicolumn{2}{c|}{Comparing} & \multicolumn{2}{c|}{Enumerating} & \multicolumn{2}{c|}{Overall} \\ \hline
Method & Acc.  & Len. & Acc. & Len. & Acc. & Len. & Acc. & Len. & Acc. & Len. \\
\hline
\hline Ours + GT Scene Graph & $92.8 \%$ & N/A & $89.2 \%$ & N/A & $93.2 \%$ & N/A & $86.0 \%$ & N/A & $90.3 \%$ & N/A \\
\hline Ours + FS + GT Segm & $81.6 \%$ & 87.4 & $63.6 \%$ & 126.5 & $83.2 \%$ & 132.5 & $51.6 \%$ & 120.8 & $70.0 \%$ & 116.8 \\
\hline Ours + FS & $60.8 \%$ & 122.5 & $34.0 \%$ & 120.3 & $62.4 \%$ & 143.9 & $10.8 \%$ & 137.8 & $42.0 \%$ & 131.1 \\
\hline Ours + GT Segm & $76.0 \%$ & 39.7 & $63.2 \%$ & 86.2 & $81.2 \%$ & 99.5 & $49.6 \%$ & 90.2 & $67.5 \%$ & 78.9 \\
\hline Ours & $58.8 \%$ & 58.0 & $31.6 \%$ & 87.5 & $62.0 \%$ & 97.7 & $12.8 \%$ & 87.8 & $41.3 \%$ & 82.7 \\
\hline
\end{tabular}
\end{center}
\caption{\textbf{K-EQA} experiment results with \textbf{ablation analysis}. ``Acc.'' stands for ``Accuracy''. ``Len.'' stands for ``Length''. }
\label{table:Ablation_Study}
\end{table*}

\section{Experiment}

In the following sections, we will report extensive experiments on the proposed dataset, including performance analysis with other methods, the ablation study on the components of our proposed framework, and the performance of the proposed framework on multi-turn and multi-agent scenarios. For all experiments without multi-turn or multi-agent scenarios, we only use the first question of the 10 questions for each test scene, which is similar to the common settings of other EQA tasks \cite{das2018embodied, yu2019multi}. The agents will be spawned at a fixed location defined in the AI2Thor. Following \cite{gordon2018iqa}, we use the overall accuracy and the accuracy of each sub-type of problems as the metric in our experiments.

\subsection{Performance Comparison}

Firstly we test the proposed method on \textbf{K-EQA} with a set of baseline methods. \textbf{Priori} guesses the most likely answer of the question only from its type according to the answer distribution. \textbf{Blindfold} is a blindfold baseline similar to \cite{anand2018blindfold}, using a Seq2Seq model similar to our proposed method but tries to predict the answer directly from the question. \textbf{IQA} uses the proposed model in IQA \cite{gordon2018iqa}, where the ground truth semantic map is used as the input of the model. \textbf{w/o KG} uses a similar approach to our proposed method, but without the knowledge base. Here the Text-to-SQL translator tries to generate a program that only looks up the scene graph.

Results of these baseline methods are reported in Table \ref{table:Baseline}. All Question-only baselines (\textbf{Priori} and \textbf{Blindfold}) perform significantly worse than our proposed method, with even LSTM used to extract information from the questions, indicating that the proposed method for balancing the dataset can mostly eliminate biases in the question. Another two baselines (\textbf{IQA}  and \textbf{w/o KG}) only look up to the question and the scene semantic map/scene graph also perform similar to the \textbf{Priori} baseline, showing the importance of knowledge base for answering questions in the proposed dataset.

\subsection{Ablation Analysis}
We perform a set of ablation studies to investigate the impact of the performance of each component of the proposed method, here are the experiment setups we adapt in the ablation study: \textbf{GT Scene Graph} uses ground truth scene graph. In this oracle setup, no exploration is performed. \textbf{GT Segm} is similar to our proposed method, but ground truth semantic segmentation is used. In the \textbf{FS} (Full Scan) setup, the planner will not use the generated program $\mathcal{P}_{plan}$ for planning to select relevant regions to search. Instead, it will scan the whole room. The result of the ablation study is reported in Table \ref{table:Ablation_Study}, from which we make the following observations:


\textbf{The proposed planner helps to reduce unnecessary exploration.} With the proposed planner, the path length is only a half to the path length of using the full scan approach, while the accuracy only drops ~1\%, showing a better trade-off between track-length and exploration accuracy.

\textbf{The text-to-SQL neural program synthesizer works fine.} When using the program generated by the text-to-SQL translator, the model reaches a high accuracy of 90.3\%, indicating that using neural program synthesis to perform joint reasoning is a viable approach.

\textbf{Scene graph generation has much room to improve.} The result of the proposed model has much room to improve according to Table \ref{table:Ablation_Study}. To investigate it, we replace the trained Mask R-CNN \cite{he2017mask} model with the ground truth segmentation (\textbf{GT Segm}). Thus, the only problem with this setup is the voxel-based scene graph generation. The model still performs well for the ``Existence'' and ``Comparing'' problem. However, for the Counting and Enumerating problems, the performance drops more significantly, indicating the voxel clustering approach could sometimes generate wrong object counts. The performance drop from ``Ours + GT Segm'' to ``Ours'' is mostly caused by instance segmentation models' imperfectness. Under the current implementation of the framework, detections of the wrong category with high confidence could be very hard to fix. Furthermore, since the asked objects are mostly small, this will make the detection more difficult.

\subsection{Qualitative Example}
\begin{figure}[h]
\begin{center}
\includegraphics[width=\linewidth]{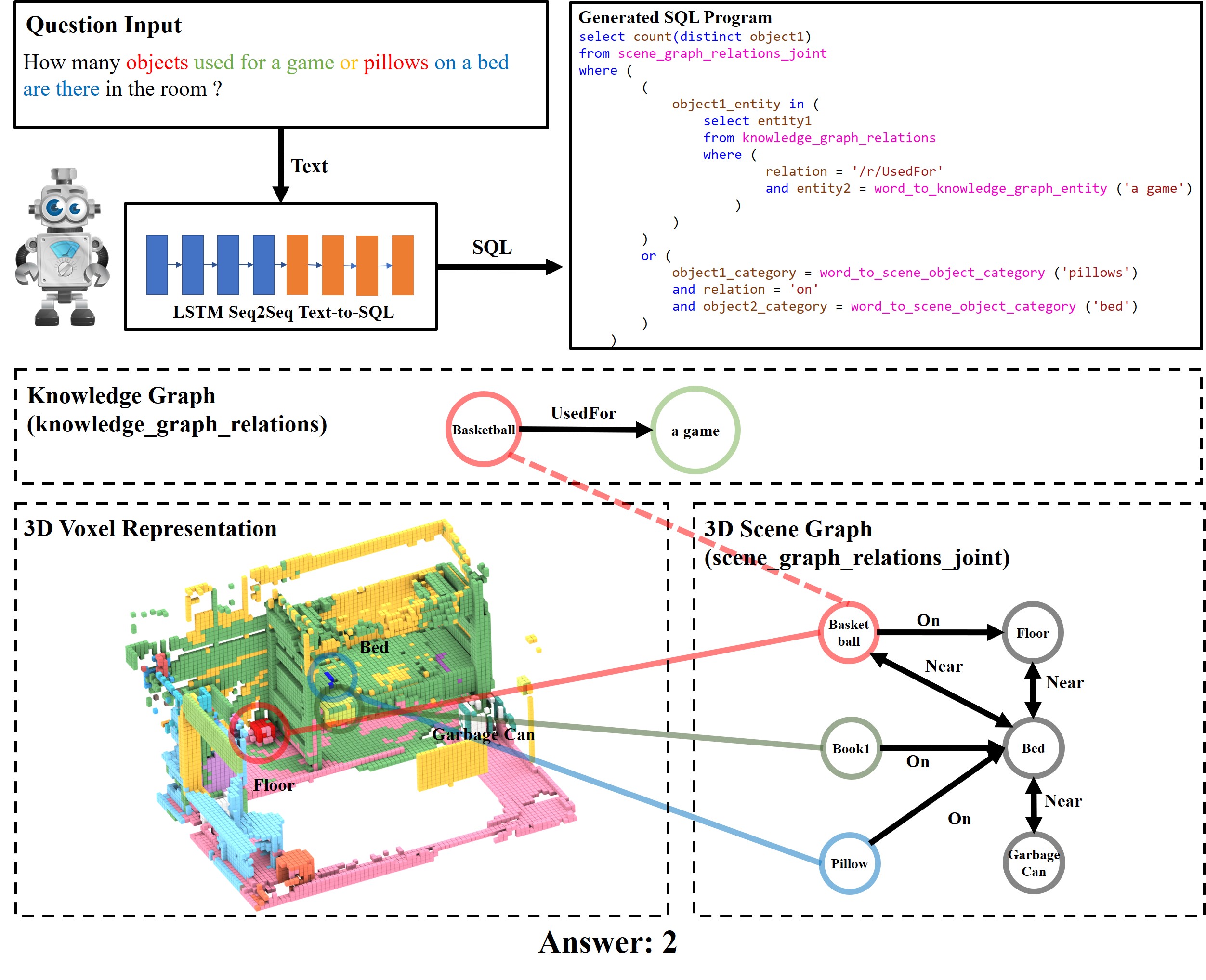}
\end{center}
   \caption{A representative example of the proposed method on the K-EQA dataset. The agent successfully explores and performs voxel-level reconstruction of the scene, and then generates a scene graph. The generated program is executed on the scene graph and the correct answer is obtained.}
\label{fig:Qualitative_Example}
\end{figure}
In Figure \ref{fig:Qualitative_Example} we present an example of the actual execution of the proposed framework. The agent first builds a voxel-level reconstruction and then uses the clustering to build the 3D Scene Graph. The generated program parses the words in the question, extracts (``used for'', ``a game'') to a ConceptNet entity and then looks it up, finds the category ``Basketball'', and finally queries all basketballs and pillows on the bed in the scene, generating the correct answer ``2''.

\subsection{K-EQA Extension Experiments}
To investigate the limit of the proposed method for questions with complex logical reasoning. We evaluate it on the proposed dataset on both \textbf{K-EQA} split and \textbf{K-EQA Extension} split. As is shown in Table \ref{table:Logic_vs_Basic}, the proposed method performs worse on the K-EQA Extension split than the K-EQA split. This indicating that with more logic connectors in the questions, the problem is becoming more challenging. The proposed framework could work on these complex questions, but will still suffer from a drop in performance.

\subsection{Multi-agent and Multi-turn Experiments}

\begin{table}[h]
\begin{center}
\begin{tabular}{|c|c|c|c|}
\hline
No. of Agents & Accuracy & Length & Speedup \\ \hline
1 & $41.3\%$ & 82.7 & 1.00 \\ \hline
2 & $41.3\%$ & 51.1 & 1.62 \\ \hline
3 & $42.2 \%$ & 42.4 & 1.95 \\ \hline
4 & $40.2 \%$ & 38.1 & 2.17 \\ \hline
5 & $40.4 \%$ & 35.6 & 2.32 \\ \hline
\end{tabular}
\end{center}
\caption{Multi-agent experiment results on \textbf{K-EQA} dataset.}
\label{table:Multi-Agent}
\end{table}


\begin{table}[h]
\begin{center}
\begin{tabular}{|c|c|c|c|}
\hline
No. of Turns & Accuracy & Length & Speedup \\ \hline
1 & $41.3\%$ & 82.7 & 1.00 \\ \hline
2 & $41.8\%$ & 51.1 & 1.62 \\ \hline
3 & $42.8\%$ & 37.5 & 2.21 \\ \hline
4 & $42.6\%$ & 29.3 & 2.82 \\ \hline
5 & $42.4\%$ & 23.8 & 3.46 \\ \hline
10 & $42.7\%$ & 12.5 & 6.64 \\ \hline
\end{tabular}
\end{center}
\caption{Multi-turn experiment results on \textbf{K-EQA} dataset.}
\label{table:Multi-Turn}
\end{table}





Our proposed framework also supports multi-turn and multi-agent question answering tasks. To demonstrate and test our method on these scenarios, the following experiment setups are also investigated:

\textbf{Multi-Turn Setup.} In this setup, the agent will be asked 10 questions in turn for one scene. The agent will get the next question only after it answers the previous one and stops exploration. This setup requires the agent to make use of its memories when answering previous questions.

Table \ref{table:Multi-Turn} shows that reconstruction and scene graph-based memory is useful for a persistent agent. Our proposed framework can keep a state memory that could fully represent the semantic structure of the scanned parts of the scene. Therefore, when an agent needs to answer multiple questions in turn, the proposed method can significantly reduce the average track length for answering each question.

\textbf{Multi-Agent Setup.} In this setup, multiple agents will be spawned at the same time, allowing them to explore and build the 3D scene graph collaboratively. All agents will be spawned at the same location as the single-agent setup. To evaluate the multi-agent efficiency, the maximum action sequence length among all agents is also computed \cite{das2019tarmac, tanmulti}.

Multi-agent experiment results in Table \ref{table:Multi-Agent} show that the proposed method could schedule multiple agents effectively. The average maximum step of all agents declines as the number of agents increases. In this experiment setup, the best speed-up ratio we can achieve is 2.32 for 5 agents.

\section{Conclusion}
We present Knowledge-based Embodied Question Answering --- a new task where the agent answers questions based on external knowledge and scene graph in 3D embodied environment. We develop a new dataset containing complex questions with logical clauses and knowledge-related phrases for this task. To address this, we propose a novel framework based on neural program synthesis, where joint reasoning of the external knowledge and 3D scene graph is performed to realize navigation and question answering. Experiments show that our framework could answer more complicated questions requiring knowledge and logical reasoning. It can also support multi-agent and multi-turn scenarios with the help of the scene graph. For future works, a stronger scene graph generation model (\eg a point-cloud-based model) can further improve the performance of the task.


%

\ifCLASSOPTIONcompsoc
  \section*{Acknowledgments}
\else
  \section*{Acknowledgment}
\fi

This work was supported in part by the National Natural Science Foundation of China under Grants 62025304.

\ifCLASSOPTIONcaptionsoff
  \newpage
\fi



%
{\small
\bibliographystyle{IEEEtran}
\bibliography{egbib}
}

%








\end{document}